\documentclass[letterpaper, preprint, paper,11pt]{AAS}	

\usepackage{bm}
\usepackage{amsmath}
\usepackage{amsfonts}
\usepackage[colorlinks=true, pdfstartview=FitV, linkcolor=black, citecolor= black, urlcolor= black]{hyperref}
\usepackage{overcite}
\usepackage{footnpag}			      	

\usepackage{nameref}  
\usepackage{subcaption}
\usepackage{makecell}
\usepackage{float} 

\usepackage{longtable}
\usepackage{graphicx}    
\usepackage{booktabs} 
\usepackage{ifthen}

\PaperNumber{25-848}

\newboolean{comparemode}
\setboolean{comparemode}{false}

\newcommand{\hr}[1]{%
  \ifthenelse{\boolean{comparemode}}%
    {\textcolor{red}{#1}}%
    {#1}%
}

\begin{document}

\title{Fast Learning of Non-Cooperative Spacecraft 3D Models through Primitive Initialization}

\author{Pol Francesch Huc\thanks{PhD Candidate, Department of Aeronautics and Astronautics, Stanford University, Stanford, CA 94305},  
Emily Bates\footnotemark[1],
\ and Simone D'Amico\thanks{Associate Professor, Department of Aeronautics and Astronautics, Stanford University, Stanford, CA 94305.}
}

\maketitle{}

\begin{abstract}
The advent of novel view synthesis techniques such as NeRF and 3D Gaussian Splatting (3DGS) has enabled learning precise 3D models only from posed monocular images. 
Although these methods are attractive, they hold two major limitations that prevent their use in space applications: they require poses during training, and have high computational cost at training and inference. 
To address these limitations, this work contributes: (1) a Convolutional Neural Network (CNN) based primitive initializer for 3DGS using monocular images; (2) a pipeline capable of training with noisy or implicit pose estimates; and (3) and analysis of initialization variants that reduce the training cost of precise 3D models. 
A CNN takes a single image as input and outputs a coarse 3D model represented as an assembly of primitives, along with the target's pose relative to the camera. 
This assembly of primitives is then used to initialize 3DGS, significantly reducing the number of training iterations and input images needed -- by at least an order of magnitude. 
For additional flexibility, the CNN component has multiple variants with different pose estimation techniques. 
This work performs a comparison between these variants, evaluating their effectiveness for downstream 3DGS training under noisy or implicit pose estimates. 
The results demonstrate that even with imperfect pose supervision, the pipeline is able to learn high-fidelity 3D representations, opening the door for the use of novel view synthesis in space applications. 
\end{abstract}

\section{Introduction and Motivation}
Recent years have seen an increase in the number of space missions that require autonomous Rendezvous, Proximity Operations, and Docking (RPOD). 
In some cases, satellites operating near another defunct or adversary satellite must perform uncooperative navigation with respect to an unknown body. 
Additionally, if the other body is unknown, then a mission may also need to reconstruct the 3D shape of the target. 
Vision-based methods for shape estimation and reconstruction use onboard cameras, which are relatively cheap, space proven, and lightweight.
Methods using monocular camera imagery are particularly valuable: in the event that stereo imagery is unavailable due to cost or equipment failure, a single camera can still be used for navigation and characterization.

High-precision tasks such as grasping and docking will require high-fidelity shape models.
Most current methods, such as feature tracking\cite{dor_orb-slam_2018} and shape reconstruction via primitives\cite{park_rapid_2024} do not achieve sufficiently detailed 3D target models. 
Recent work leverages new advances in Novel View Synthesis (NVS) \cite{aerospace11030183}, a class of techniques that train implicit representations of scenes or targets that are then used to render unseen views of the target or determine its pose from incoming images. \cite{yen-chen_inerf_2021}
However, gaps in implementation prevent existing NVS methods from being readily deployable on space platforms.
NVS methods such as Neural Radiance Fields (NeRFs) \cite{mildenhall2020nerf} and 3D Gaussian splatting (3DGS) \cite{kerbl_3d_2023} typically require the pose (relative position and orientation between observer and target) to be known during the model's training or rely on the availability of depth measurements \cite{Barad_2024}.
Furthermore, existing NVS works for space applications do not address the high computational load of NVS algorithms, which can be challenging to deploy on a compute-limited space-grade processor.

3D Gaussian Splatting (3DGS) is an NVS algorithm that gradually trains a model of a scene constructed from 3D Gaussians. 
By training the location, size, and color of thousands of 3D Gaussians, the method is capable of producing high fidelity renders of the subject, implicitly storing a precise 3D model in the process.
In prior implementations, 3DGS requires tens to hundreds of images and thousands of training iterations to train an accurate model.
The training is also done on desktop grade GPUs, which far outclass GPUs that may be available on space missions in the near future. 
Like other NVS methods, 3DGS requires pose priors which are typically calculated in a large batch using the COLMAP structure-from-motion algorithm\cite{schoenberger2016mvs}$^,$ \cite{schoenberger2016sfm} .
This batch pose estimation approach is undesirable in precise space rendezvous applications due to the latency in the collection of the image batch.
 
This work seeks to address gaps in NVS for space applications by reducing the computational load during NVS initialization.
By accelerating the initialization of a 3D Gaussian satellite model, this work seeks to greatly reduce the number of training iterations and image inputs needed to create a high-fidelity 3D Gaussian splatting model.
Instead of the typical naive random initialization, a low-fidelity shape estimate from a neural network provides a coarse initialization to a 3DGS training pipeline.
The pose of this target is also recovered from the neural prior, enabling a full end-to-end pipeline for relative navigation and precise visual characterization of an unknown, uncooperative target using only monocular images.

This paper begins by discussing the state of the art in this field. 
Next, the proposed architecture is detailed, with detailed descriptions of both the 3DGS training process and the neural network providing the coarse initial shape estimate. 
After that, experimental results are shown on the SPE3R \cite{park_rapid_2024} dataset, and the work, impact, and potential next steps are summarized in the conclusion.

\section{State of the Art}
One of the first space missions to demonstrate autonomous vision-based cooperative rendezvous was DARPA's Orbital Express \cite{4526518} in 2007, which used a near-field camera to provide 6-DOF measurements of the target spacecraft. 
This cooperative mission relied on retro-reflectors on the target spacecraft. 
The PRISMA mission, launched in 2011, used classical computer vision techniques to regress the 6D pose of a known target without retro-reflectors \cite{d_pose_2014}, making it an early demonstration of uncooperative rendezvous. 
The mission relied on classical image processing techniques to detect features in the target which were then matched against the known 3D model. 
To improve the accuracy of the pose estimates, deep Convolutional Neural Networks (CNN) \cite{sharma_pose_2018} have been implemented which are trained on a dataset of images of the known target. 
More recently, CNNs have been combined with classical estimation techniques \cite{park_adaptive_2023} in order to further refine the pose estimates and improve the robustness to the harsh lighting conditions found in space. 
The flight readiness of these neural-network-based pose estimators has also been improved through the use of online model refinement \cite{park2024spnv2} and the adoption of vision transformers \cite{park2024spnv3}.
However, these methods require previous knowledge of the target to generate the dataset and train the neural network. 


The above machine learning methods, which are trained on large amounts of images, have lead to the release of different publicly available image datasets of satellites. 
Early examples of these datasets include URSO \cite{9197244} and Dung et al. \cite{Dung_2021_CVPR} which include both synthetic and space images of multiple targets but do not have pose labels for any or all images. Datasets such as SPEED+ \cite{park2021speedplusdataset} SHIRT \cite{park_adaptive_2023}, and MAN-DATA \cite{lebreton_training_2024} on the other hand, include synthetic and real images in a space-surrogate environment and provide reference truth poses for all images, providing a source of validation. Finally, SPE3R \cite{park_rapid_2024} provides synthetic images of 64 real satellites with the associated pose labels. 

Work on high-fidelity shape reconstruction for unknown satellites has been comparably limited.
Park et al. \cite{park_rapid_2024} introduced the SPE3R dataset and trained a CNN on this dataset to estimate the pose and a coarse shape as a set of primitive assemblies from a single image. 
However, the CNN suffers from lack of generalization, showing degraded performance when tested with previously unseen satellites. 
In a follow-on paper, Park et al. \cite{park_improving_2024} attempted to improve the shape estimates by extending the CNN through better sampling of the surface, the application of a transformer for auto-regressive inference, and the use of part labels of the target for additional supervision, but still showed a large gap in performance between the training and testing sets. 
Bates et al. \cite{bates_removing_2025} on the other hand, focused on improving the CNN's pose estimates by either handling ambiguities caused by a lack of canonical pose in satellites, or outright removing them by defining the target's body axes to be parallel to the camera axes. 
CRISP \cite{shi_crisp_2024} follows a similar approach of using a low parameter representation of the 3D model by using a Signed Distance Field (SDF), and trains a neural network capable of estimating the pose and shape from a single RGB-D image. As noted, CRISP relies on RGB-D images which have depth measurements that would require an additional sensor and hence be less robust than methods that only rely on RGB images.

Due to the low number of parameters required to describe the primitive assemblies introduced in Park et al. \cite{park_rapid_2024} or the SDF in CRISP \cite{shi_crisp_2024}, they are well suited for small neural networks, but they do not provide a fine 3D model of the target. 
To increase the precision of the 3D model while sacrificing some computational load, others represent their targets using point-clouds.
ORB-SLAM, a common terrestrial algorithm, was successfully applied to space imagery \cite{dor_orb-slam_2018}, but still only provided a coarse 3D model. 
Structure from motion has also been applied to the reconstruction problem, but previous works such as Dennison et al. \cite{dennison_leveraging_2023} have made assumptions about the available information such as knowing the camera orientation with respect to the target, thus reducing their applicability. 

NVS methods offer even higher 3D model precision, at the cost of computational load.
Nguyen et al. \cite{aerospace11030183} show that 3DGS can be applied to space-like images and analyzed the computational costs on a consumer grade GPU, but did not reduce computation and still relied on batch pose estimation.
Mathihallia et al. \cite{mathihalli2024dreamsatgeneral3dmodel} successfully fine-tune an existing model which allows the generation of shape estimates from a single image that are more precise than the assembly of primitives method described in Park et al. \cite{park_rapid_2024}, but unlike Park et al. \cite{park_rapid_2024} they do not estimate the pose. 
Barad et al. \cite{Barad_2024} also represent the 3D model using 3D gaussian splats and accurately estimate the pose estimates for some time before the lighting changes and the images no longer appear similar. Like CRISP though, this method relies on RGB-D images which are less robust than methods that rely on RGB images.

Although the works discussed above present a solid foundation for unknown target pose estimation and shape reconstruction, there is no single implementation that simultaneously achieves (a) self-contained monocular shape and pose estimation without additional sensor data, (b) high-precision shape estimation suitable for docking or similar procedures, and (c) reduced computational requirements that are suitable for a spaceborne processor.
Park et al.'s CNN \cite{park_rapid_2024} meets requirements (a) and (c), so it has been selected for use as an initialization method to be paired with a 3DGS training pipeline.

\section{Methodology}

The proposed architecture consists of two main parts: the formation of an initial shape estimate and the training of a 3DGS model which is initialized using this shape estimate.
To form the initial estimate, an initial image is passed to a CNN which generates an initial shape and pose estimate.
Then, points from the surface of the initial shape model are sampled and used to initialize the 3D Gaussians.
This collection of Gaussians is passed to the 3DGS training pipeline, which uses successive incoming images to train a 3DGS representation of the target supervised by corresponding pose estimates from the CNN.
The entire training pipeline is shown in Figure \ref{fig:pipeline}, and detailed descriptions of each step are given below.

\begin{figure}[t]
    \centering
    \includegraphics[width=\textwidth]{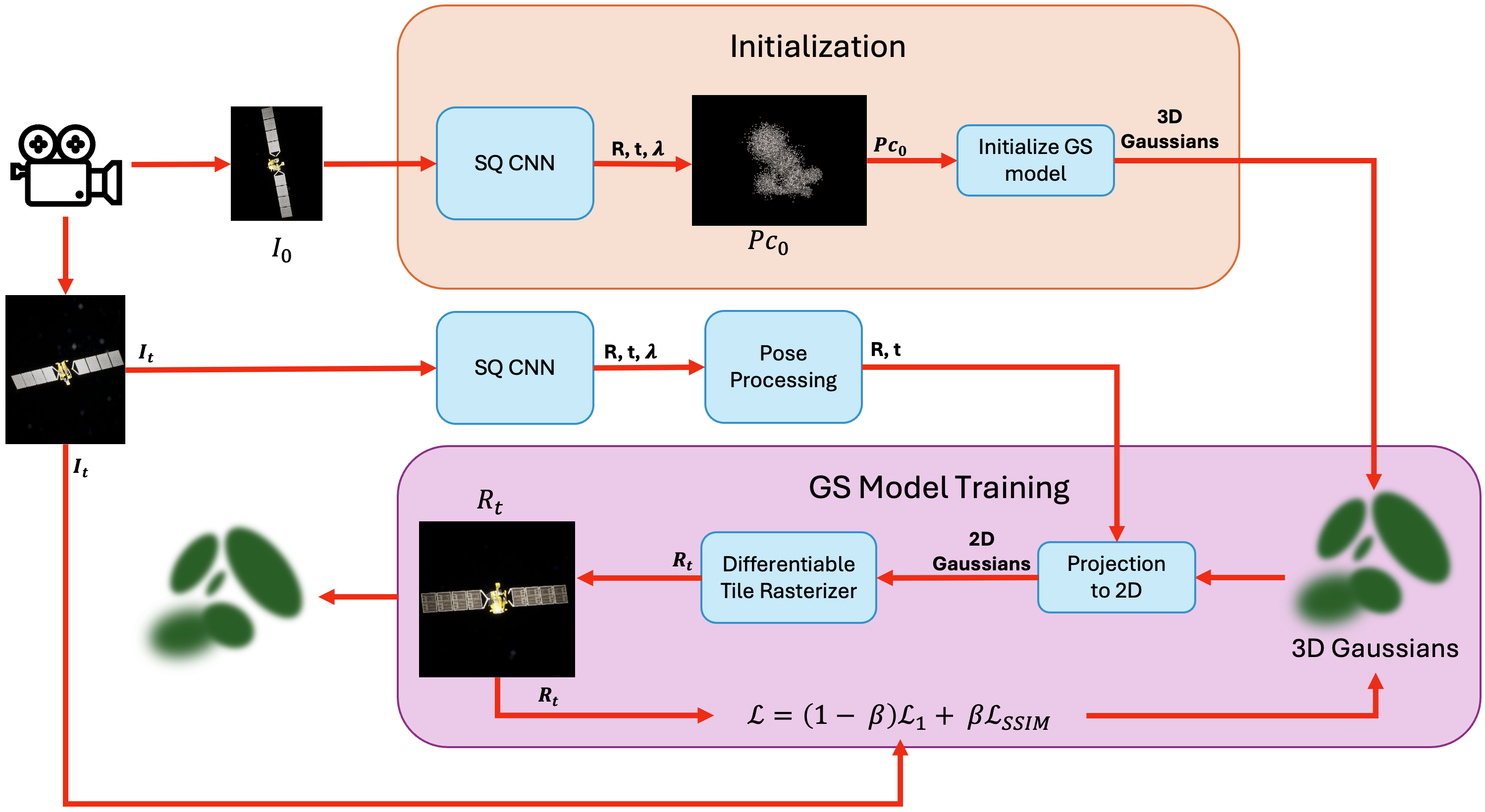}
    \caption{Pipeline for accelerated 3DGS model training using shape and pose estimation CNN. The initial shape model is created in the ``Initialization" portion where the CNN outputs rotation $R$, translation $t$, and shape $\lambda$ estimates based on the image. The shape estimate is then used to sample a point-cloud, which is then used to jumpstart training in the ``3DGS Model Training" portion.}
    \label{fig:pipeline}
\end{figure}

\subsection{Estimating a satellite as an assembly of primitives}\label{sec:cnn_estimation_primitives}
This work uses the CNN introduced by Park et al. \cite{park_rapid_2024} and extended by Bates et al. \cite{bates_removing_2025}.
The network takes in a single image of a close-range spaceborne target and returns both an estimate of the target's shape and an estimate of its pose (translation and rotation with respect to the camera axis frame).

The target's shape is represented as an assembly of primitives known as superquadrics.
Superquadrics are represented by a relatively small number of parameters, enabling them to be easily learned by a neural network.
The $i^{th}$ primitive is parameterized by 
\begin{equation}
    \boldsymbol{\lambda}_i =   [ \boldsymbol{\alpha}_i, \boldsymbol{\epsilon}_i, \mathbf{t}_i, \mathbf{r}_i, \mathbf{k}_i ] 
\end{equation}
where $\boldsymbol{\alpha}_i \in \mathbb{R}^3$ represents the primitive's shape, $\boldsymbol{\epsilon}_i \in \mathbb{R}^2$ represents the primitive's size, $\mathbf{t}_i \in \mathbb{R}^3$ gives its translation from the target's body frame origin, $\mathbf{r}_i \in \mathbb{R}^6$ gives its rotation from the target body axis frame as a 6D rotation vector, and $\mathbf{k}_i \in \mathbb{R}^2$ describes tapering along the primitive's x and y directions. 
Hence, the entire shape model may be described by a total of 16$\times$$M$ parameters, where the model consists of $M$ primitives.
A set of analytical equations retrieve points sampled from the surface and interior of the primitive using these parameters, as described by Park et al.\cite{park_rapid_2024}

The target's estimated pose consists of the relative translation $\widetilde{\mathbf{t}}$ and the relative rotation $\widetilde{\mathbf{R}}$.
The relative translation is the translation between the camera origin and the target's body frame origin. 
The relative rotation is the rotation between the camera's body frame and the target body frame.
The CNN outputs estimated rotation as a 6D rotation vector because of its suitability for training in neural networks.\cite{zhou_continuity_2020} 
This vector can be easily  converted to a 3$\times$3 direction cosine matrix for later use.

The network is trained using the SPE3R dataset \cite{park_rapid_2024} with a combination of translation, rotation, reprojection, and chamfer losses.
The translation and rotation losses evaluate how close the pose estimate is to the true pose. 
Reprojection loss evaluates how well the image can be reconstructed using the estimated shape and pose.
Chamfer loss evaluates the quality of the 3D shape reconstruction.
The mathematical formulations for these losses are detailed in previous work.\cite{park_rapid_2024}$^,$\cite{bates_removing_2025}

Two new formulations of the CNN were introduced in 2025 to account for ambiguities in the shape and pose representations.\cite{bates_removing_2025}
An ambiguity-aware implementation trains the network to align the shape model based on its principal dimensions without penalizing for ambiguities in this alignment; it provides the outputs described above, where the shape and rotation estimate may be permuted along the shape model's principal dimensions.
For instance, a single rotation estimate from the CNN is permuted during training by
\begin{equation}
    \widetilde{\mathbf{R}}_{\text{P}} \in \widetilde{\mathbf{R}} \cdot \left\{ \begin{bmatrix}
        1 & 0 & 0 \\
        0 & 1 & 0 \\
        0 & 0 & 1
    \end{bmatrix}, \begin{bmatrix}
        1 & 0 & 0 \\
        0 & -1 & 0 \\
        0 & 0 & -1
    \end{bmatrix}, \begin{bmatrix}
        -1 & 0 & 0 \\
        0 & 1 & 0 \\
        0 & 0 & -1
    \end{bmatrix}, \begin{bmatrix}
        -1 & 0 & 0 \\
        0 & -1 & 0 \\
        0 & 0 & 1
    \end{bmatrix} \right\} \label{eq:rot_perm}
\end{equation}
Training losses are computed for each permutation and the lowest scoring loss is applied during training.
This helps to avoid unfair training penalties due to a different alignment of the estimated shape versus the ground truth shape model.
This variant is therefore able to generate a pose and shape combination that effectively reproduces the given scene, even when it has not been trained on images of the given satellite.

An ambiguity-free implementation removes pose and shape ambiguities by defining the target body axis frame parallel to the camera axis frame and allowing the shape estimate to represent the instantaneous configuration of the target at the time of imaging.
Hence, a rotation estimate is not needed as the rotation matrix in this case is a 3$\times$3 identity matrix.

\subsection{3D Gaussian Splatting} \label{sec:vanilla_3dgs}
3D Gaussian Splatting (3DGS) \cite{kerbl_3d_2023} is the novel view synthesis method used in this work. 
When trained, it is able to render high fidelity scenes or objects from arbitrary viewpoints, including those not seen during training. 
The scene or object is represented by a collection of 3D Gaussians $\mathcal{G}$ which are scattered in the world and are parameterized by a location $\mu$, covariance $\sigma$ as
\begin{align}
    G(x) = \frac{1}{2\pi \sqrt{|\Sigma}|}e^{-\frac{1}{2}(x-\mu)^T \Sigma^{-1}(x-\mu)}
\end{align}
and additionally have an appearance set by the opacity $\alpha$, and spherical harmonics to encode view-dependent color $c$. 
To render an image, the 3D Gaussians $\mathcal{N}(\mu, \Sigma)$ are first projected to the image plane through a perspective projection $\pi$ into 2D Gaussians $\mathcal{N}(\mu_I, \Sigma_I)$
\begin{align}
    \mu_I = \pi(T_{CW} \mu), \quad \Sigma_I = J R \Sigma R^T J^T
\end{align}
where $T_{CW}$ is the transform from the camera to the world reference frame. $J$ is the Jacobian of the approximation of the projective transformation and $R$ is the rotation component of $T_{CW}$. 

This projection splats the 3D Gaussians onto the image plane, where they can be sorted front to back.
Each pixel color value $C_p$ must be synthesized individually by summing over the individual Gaussians $N$ which are in front of this pixel:
\begin{align}
    C_p = \sum_{i \in N} c_i \alpha_i \prod_{j=1}^{i-1}(1- \alpha_j)
\end{align}
During training, successful Gaussians are split and cloned so that regions of interest can be more detailed. 
On the other hand, Gaussians which are too transparent to ever be seen are culled to reduce the computational cost. 

The training of this novel view synthesis method is fully differentiable, allowing for it to be incrementally trained by comparing a real image taken at the pose with the rendering from 3DGS. 
The loss function for this method combines the $\mathcal{L}_1$ loss function with a Structural Similarity Index Measure (SSIM) \cite{1284395} term:
\begin{equation}
    \mathcal{L} = (1 - \beta) \mathcal{L}_1 + \beta \mathcal{L}_{SSIM} \label{eqn:3dgs_full_loss}
\end{equation}
where $\beta$ is a hyper-parameter to balance the two losses. Mathematical definitions of these loss functions are provided in the section ``\nameref{sec:eval_metrics}".
Unlike more typical 3D modeling approaches like meshes or voxels, 3DGS does not store the model of the subject explicitly. 
Instead the scene or object geometry and appearance are stored implicitly by the parameters of the 3D Gaussians which allows for efficient high quality rendering. 


The original 3DGS implementation by Kerbl et al. \cite{kerbl_3d_2023} is designed for offline reconstruction using a large batch of images.
However, in space-based applications image acquisition is typically slow -- when compared with terrestrial applications -- making it impractical to wait for a complete set of images before beginning reconstruction. 
To address this, the 3DGS pipeline is adapted to operate in a sequential manner, enabling the model to update incrementally as new images become available. 
In the implementation used here, each new image is used to refine the reconstruction for a fixed number of training steps before being discarded, after which the system waits for additional images. 
This sequential reconstruction strategy allows the model to improve over time with minimal delay, reducing the total time required to obtain an accurate 3D model of the target.

\subsection{Initializing 3DGS from primitives} \label{sec:initialize_3dgs}

\begin{figure}[ht]
    \captionsetup[subfigure]{labelformat=empty} 
    \centering
    \begin{subfigure}{0.24\textwidth}
        \includegraphics[width=\textwidth]{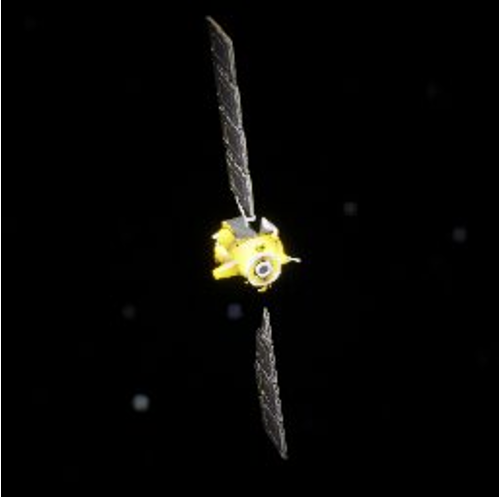}
    \end{subfigure}
    \begin{subfigure}{0.24\textwidth}
        \includegraphics[width=\textwidth]{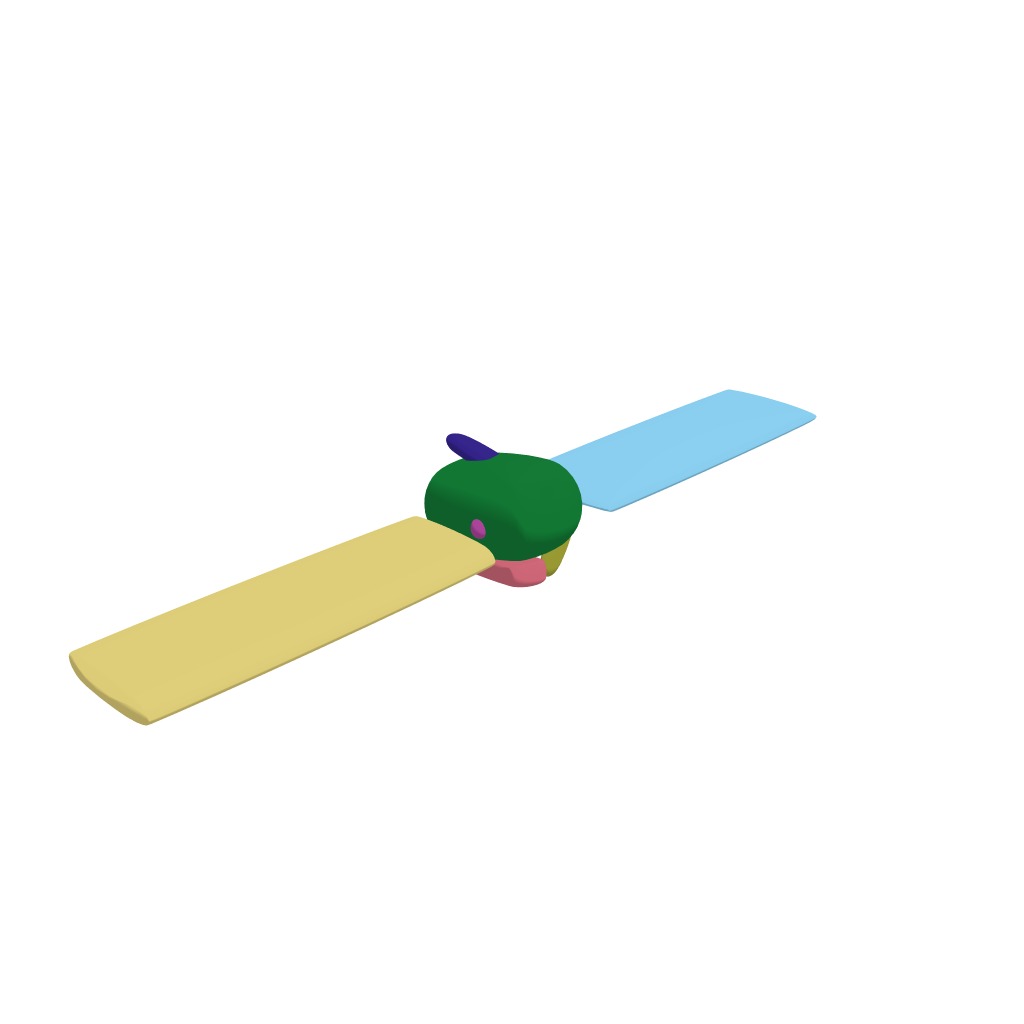}
    \end{subfigure}
    \begin{subfigure}{0.24\textwidth}
        \includegraphics[width=\textwidth]{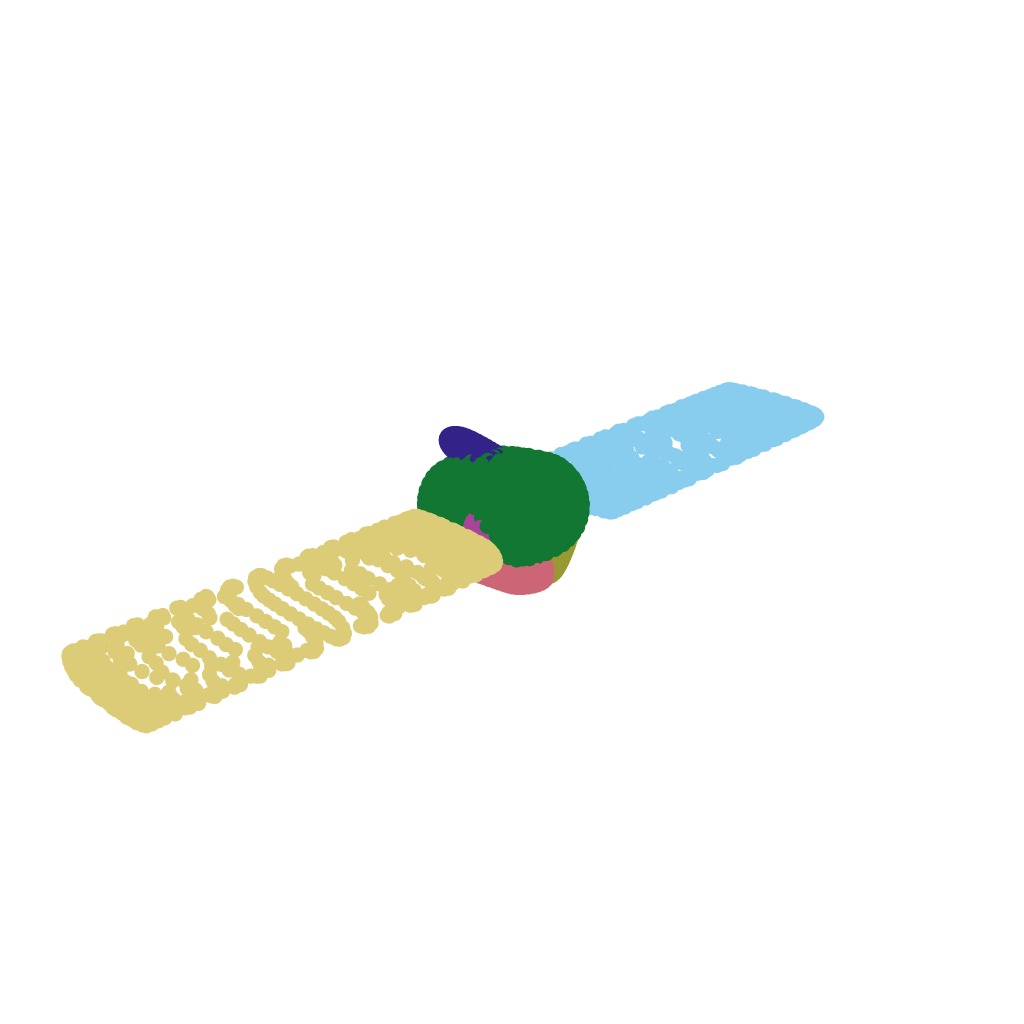}
    \end{subfigure}
    \begin{subfigure}{0.24\textwidth}
        \includegraphics[width=\textwidth]{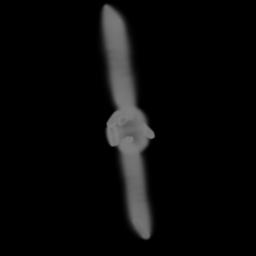}
    \end{subfigure}
    \caption{Example of proposed 3DGS model initialization. Given an input image (far left), the CNN generates an initial shape estimate as a set of primitives (center left). 3D points are sampled from the shape model's surface (center right) and used to create the initial 3DGS model, shown at far right along the original view angle. }
    \label{fig:image2primitives2pointcloud}
\end{figure}

The 3D Gaussians used in 3DGS can be thought of as a natural extension to point clouds. 
Hence, to initialize from an assembly of primitives, the proposed approach samples the surface of these primitives to generate a set of 3D points which are then used as mean positions for the initial set of Gaussians. 
This process is visualized in
\autoref{fig:image2primitives2pointcloud}, showing the process of converting an initial shape estimate (as a primitive assembly) to the initial set of Gaussians.

\subsection{Source of pose}

Once 3DGS is initialized, the model is trained by comparing renderings at a given pose to the true images collected, as discussed in section ``\nameref{sec:vanilla_3dgs}". 
To resolve the need for pose, this work leverages the pose estimates from the CNN and uses them as the reference truth values when training 3DGS. The original pose and shape estimation CNN from Park et al.\cite{park_rapid_2024} and the ambiguity-aware variant introduced by Bates et al.\cite{bates_removing_2025} directly provide pose estimates.

The ambiguity-free CNN variant \cite{bates_removing_2025} does not explicitly estimate the rotation of the target with respect to the camera and instead estimates the shape of the target with respect to a body axis frame defined parallel to the camera axes. 
In theory this means that the rotation provided is exact, as the CNN effectively estimates the rotation matrix as the identity.
In practice for this work there is a desire to use explicit pose labels with respect to a rigid body, so the rotation of the target must be extracted from the model.
To this end, point cloud alignment is performed between the point-cloud from the first primitive assembly estimated by the CNN $S_0$ and the current one $S_i$.
Thus, the body axis of the target is established as identity at the time of the first image, and subsequent poses indicate its rotation with respect to this initial pose.
Since the ambiguity-free CNN variant still estimates the position of the target with respect to the camera, point-cloud alignment is used to estimate the rotation only.
This is done by minimizing the chamfer distance $CD$:
\begin{align}
    CD(S_0, \tilde{S}_i) &= \frac{1}{|S_0|} \sum_{x \in S_0} \min_{y\in \tilde{S}_i} ||x - y||_2^2 + \frac{1}{|\tilde{S}_i|} \sum_{y\in \tilde{S}_i} \min_{x \in S_0} ||x - y||_2^2 \label{eqn:chamfer_distance}
\end{align} 
where $\tilde{S}_i$ refers to the point-cloud at the current evaluation of the CNN rotated by the rotation estimate $R_i$
\begin{align}
    \tilde{S}_i = S_i R_i
\end{align}

Under a sequential data stream where pose estimates are available through a sequential filter, one would initialize the optimization with the rotation estimate from the filter. 
Point-cloud alignment would then be used to refine the rotation estimate by minimizing the chamfer distance.
In the case of datasets where the images are randomly distributed, no such pose estimate exists to be refined.
The simple solution is to initialize the point-cloud alignment with an identity rotation, but this can quickly lead to local minima.
Instead, a sampling approach is taken where 16 quasi-random samples of initial orientation estimates are selected, optimized by minimizing the chamfer distance, and the one with the lowest chamfer distance is used as the pose estimate. 
This method can still lead to local minima, but experiments demonstrate that this is a good balance between computational cost and accuracy.

\subsection{Evaluation metrics} \label{sec:eval_metrics}
This work explores two sets of evaluation metrics that are commonly used in Novel View Synthesis (NVS) methods for training and evaluation.
The first set is used during training of the 3DGS model, and as shown in \autoref{eqn:3dgs_full_loss} is minimized through gradient descent.
The first metric in \autoref{eqn:3dgs_full_loss} is the $\mathcal{L}_1$ image loss, 
\begin{equation}
    \mathcal{L}_1 = \frac{1}{N} \sum_{i=1}^N || R_t^i - I_t^i||_1 \label{eqn:l1_image_loss}
\end{equation}
where $R_t^i$ refers to the pixel location $i$ of the rendering at time $t$, and $I_t^i$ of the collected image. 
This loss then performs a pixel-wise comparison between the rendering and image, and guides the rendering towards a certain visualization of the subject during the training. 
The second metric used is the Structural Similarity Index Measure (SSIM) \cite{1284395} which compares the edges and textures between two images. 
It is calculated from the mean and variance of the images
\begin{equation}
    SSIM(I, R) = \frac{(2 \mu_I \mu_R + C_1)(2 \sigma_{IR} + C_2)}{(\mu_I^2 + \mu_R^2 + C_1)(\sigma_I^2 + \sigma_R^2 + C_2)}
\end{equation}
where $\mu_I$ and $\sigma_I$ refer to the mean and variance of the image respectively, with $\mu_R$ and $\sigma_R$ for the rendering. 
$\sigma_{IR}$ is the covariance of the image and rendering, with $C_1$ and $C_2$ as stabilization constants to prevent division by 0. 
The mean, variance, and covariance for the images are calculated for a grayscale image $I \in \mathbb{R}^{H \times W}$ as:
\begin{align}
    \mu_I &= \frac{1}{N} \sum_{i=1}^H \sum_{j=1}^W I_{ij} \\
    \sigma_I^2 &=  \frac{1}{N} \sum_{i=1}^H \sum_{j=1}^W \left( I_{ij} - \mu \right)^2 \\
    \sigma_{IR}^2 &= \frac{1}{N} \sum_{i=1}^H \sum_{j=1}^W \left( I_{ij} - \mu_I \right) (R_{ij} - \mu_R)
\end{align}
where $I_{ij}$ is the pixel intensity at location $(i,j)$. $\mu_R$ and $\sigma_R$ follow the same pattern.

The next set of metrics is not used during training, but is used to provide additional context regarding the quality of the reconstruction and the 3D model. 
The first metric in this set is the Peak Signal to Noise Ratio (PSNR) which is calculated from the Mean Squared Error (MSE) as
\begin{equation}
    PSNR = 20 \log_{10} \frac{1}{\sqrt{MSE}} 
\end{equation}
where
\begin{equation}
    MSE = \frac{1}{N} \sum_i^t \left( R_t^i - I_t^i \right)^2
\end{equation}
represents the MSE with the same variables as were used in \autoref{eqn:l1_image_loss}. 
As defined, the PSNR metric is only valid for normalized images where the largest pixel intensity is $1$.
The next metric used for evaluation purposes only is the Learned Perceptual Path Similarity (LPIPS) loss, \cite{zhang2018perceptual} which compares feature activations from AlexNet for two different images.
This loss measures the perceptual difference between two images and does not focus so much on the pixel-level details. 
The final metric used in this work is the chamfer distance from \autoref{eqn:chamfer_distance}.
However, instead of comparing superquadrics-based point-clouds, here the 3D Gaussians are reduced to a point-cloud representation, which is then compared with a point-cloud sampled from the ground truth 3D model. 

\section{Results}

The results are divided into two distinct sets of experiments, each tackling the two main problems with NVS identified by this work. 
The first set demonstrates that when using the SQ CNN to initialize 3DGS, one can learn a refined 3D model much faster than with the baseline initialization method.
To showcase the full potential of this approach, these experiments use the reference truth poses.
In the following set of experiments, the estimated poses from the three CNN variants are used as the pose inputs for 3DGS training to showcase the full pipeline. 

Before describing the results, the implementation details and the hyper-parameter tuning of 3DGS are discussed below.

\subsection{Implementation Details}

As described in the section ``\nameref{sec:vanilla_3dgs}", the original 3DGS method is re-implemented such that the training does not occur over a pre-collected batch of images and is instead trained on a single image before moving onto the next one. 
The learning is performed by optimizing over the loss in \autoref{eqn:3dgs_full_loss} with the hyper-parameter $\beta = 0.2$. For every image, 5 training steps are taken and then the image is discarded for the next one in the dataset.

To initialize 3DGS, two approaches are tested: random, and by SQ CNN. The random initialization follows the implementation in the original work by Kerbl et al. \cite{kerbl_3d_2023} producing a point-cloud of 5,000 points randomly distributed that is used to initialize the model.
The latter uses the neural network implementations and trained weights provided by Park et al. \cite{park_rapid_2024} and Bates et al. \cite{bates_removing_2025} to produce shape and pose estimates. 

The method is tested on the SPE3R dataset \cite{park_rapid_2024}, which includes 64,000 synthetic images of 64 satellites -- 1,000 images per satellite -- and the accompanying masks. 
The training of the CNNs introduced by Park et al. \cite{park_rapid_2024} and continued by Bates et al. \cite{bates_removing_2025} is also conducted on this dataset, and it splits the satellites into training and testing sets.
In particular, 57 of the satellites and their associated images are used in training and validation, whereas 7 are used for testing. 
The latter set represents the true mission conditions, where the CNN is exposed to satellites it has never seen before. 
Only 300 images are used for training the 3DGS models, since that is generally sufficient to train the 3DGS model with the initialization approach proposed in this work. 
Since there are 5 training iterations per image, that means that there are only 1,500 training iterations per model. 

The training and timing metrics are computed on a desktop computer with an NVIDIA RTX 4090.
The timing metrics are therefore provided as a comparative assessment between the initialization methods and to demonstrate that the additional time spent initializing still leads to faster convergence. 

\subsection{Hyper-parameter Tuning}
The implementation of 3DGS provided by the original authors \cite{kerbl_3d_2023} includes tuned hyper-parameters. 
These have been tuned to train models on terrestrial scenes with an initialization from a COLMAP or random point-cloud, so further performance improvements can be expected from some tuning of these parameters. 
To tune the hyper-parameters, a coarse grid search is performed over a subset that is selected as the most likely to affect reconstruction accuracy and speed. 
\autoref{tab:hyperparameters_3dgs} lists the hyper-parameters which were tuned in this work compared against their values from the original 3DGS implementation. 
Readers are referred to Kerbl et. al. \cite{kerbl_3d_2023} for a full list of hyper-parameters.  

\begin{table}[ht]
    \centering
    \small
    \caption{Hyper-parameters for 3DGS training which have been changed for the two cases -- use of reference truth poses or the estimates.}
    \begin{tabular}{r|c c c}
        \toprule
        Hyper-parameter & Original & RT Poses & Estimated Poses \\
        \hline
        Initial SH degree & 0 & 2 & 1 \\
        Maximum SH degree & 3 & 3 & 3 \\
        Rate to increase SH degree & 1,000 & 1,000 & 500 \\
        Rate to prune and densify Gaussians & 100 & 100 & 100 \\
        Iteration at which to start densification and pruning & 500 & 500 & 100 \\
        Total number of training iterations & 30,000 & 1,500 & 1,500 \\
        \bottomrule
    \end{tabular}
    \label{tab:hyperparameters_3dgs}
\end{table}

\subsection{Experiments}

3DGS training using the initialization techniques described in this work is performed on the whole of the SPE3R dataset.
The results are shown in Tables \ref{tab:gt_pose_train_set_stats} and \ref{tab:gt_pose_test_set_stats}. 
These tables show the metrics used for training and evaluation at the last training iteration, averaged over the relevant set of satellites to show the generalization capability of this method. 
In addition to the training and evaluation metrics, timing metrics are provided that represent the time and number of iterations required to reach a certain rendering fidelity.
For each row, an arrow indicates whether higher ($\uparrow$) or lower ($\downarrow$) values are better, and the best value in each row is highlighted in bold.

\begin{table}[ht]
    \centering
    \footnotesize
    \caption{Training, evaluation, and timing metrics ($\mu \pm \sigma$) for the SPE3R training set using reference truth poses.}
    \begin{tabular}{r|c c c c}
        \toprule
        \makecell[r]{Init. Style / \\ Metric} & Random & \makecell{CNN \\ Original} & \makecell{CNN \\ Ambiguity Aware} & \makecell{CNN \\ Ambiguity Free} \\
        \hline
        $\mathcal{L}  \; (\downarrow)$ & $0.093 \pm 0.048$ & $\mathbf{0.084 \pm 0.048}$ & $0.086 \pm 0.049$ & $0.086 \pm 0.049$ \\
        $\mathcal{L}_1  \; (\downarrow)$ & $0.060 \pm 0.033$ & $\mathbf{0.055 \pm 0.033}$ & $0.056 \pm 0.033$ & $0.055 \pm 0.033$ \\
        SSIM $(\uparrow)$ & $0.776 \pm 0.108$ & $\mathbf{0.796 \pm 0.110}$ & $0.790 \pm 0.112$ & $0.792 \pm 0.114$ \\
        \hline
        PSNR $(\uparrow)$ & $16.504 \pm 2.677$ & $\mathbf{17.635 \pm 2.999}$ & $17.435 \pm 3.048$ & $17.549 \pm 2.999$ \\
        LPIPS $(\downarrow)$ & $0.388 \pm 0.109$ & $\mathbf{0.233 \pm 0.120}$ & $0.268 \pm 0.131$ & $0.253 \pm 0.133$ \\
        CD $(\downarrow)$ & $0.233 \pm 0.519$ & $\mathbf{0.002 \pm 0.004}$ & $0.006 \pm 0.010$ & $0.003 \pm 0.007$ \\
        \hline
        Init. Time [s] $(\downarrow)$ & $\mathbf{0.005 \pm 0.002}$ & $0.243 \pm 0.061$ & $0.228 \pm 0.059$ & $0.232 \pm 0.061$ \\
        \noalign{\vskip -1.75mm} \\
        2.0 $\times$ LPIPS\textsubscript{best} \makecell[r]{Time to [s] \\ Iters to [iters]} $(\downarrow)$ & \makecell[c]{$7.590 \pm 5.353$ \\ $968 \pm 574$} & \makecell[c]{$\mathbf{0.648 \pm 0.639}$ \\ $\mathbf{46 \pm 63}$} & \makecell[c]{$1.463 \pm 1.768$ \\ $161 \pm 247$} & \makecell[c]{$1.234 \pm 1.421$ \\ $121 \pm 169$} \\
        \noalign{\vskip -1.75mm} \\
        1.5 $\times$ LPIPS\textsubscript{best} \makecell[r]{Time to [s] \\ Iters to [iters]} $(\downarrow)$ & \makecell[c]{$9.972 \pm 4.414$ \\ $1337 \pm 393$} & \makecell[c]{$\mathbf{2.409 \pm 1.775}$ \\ $\mathbf{285 \pm 197}$} & \makecell[c]{$4.924 \pm 3.685$ \\ $616 \pm 440$} & \makecell[c]{$4.089 \pm 2.684$ \\ $511 \pm 312$} \\
        \noalign{\vskip -1.75mm} \\
        1.1 $\times$ LPIPS\textsubscript{best} \makecell[r]{Time to [s] \\ Iters to [iters]} $(\downarrow)$ & \makecell[c]{$10.626 \pm 4.168$ \\ $1423 \pm 328$} & \makecell[c]{$\mathbf{7.884 \pm 3.036}$ \\ $\mathbf{994 \pm 212}$} & \makecell[c]{$10.086 \pm 3.323$ \\ $1299 \pm 253$} & \makecell[c]{$10.092 \pm 3.384$ \\ $1293 \pm 239$} \\
        \bottomrule
    \end{tabular}
    \label{tab:gt_pose_train_set_stats}
\end{table}

\begin{figure}[p]
    \centering
    \includegraphics[width=\linewidth]{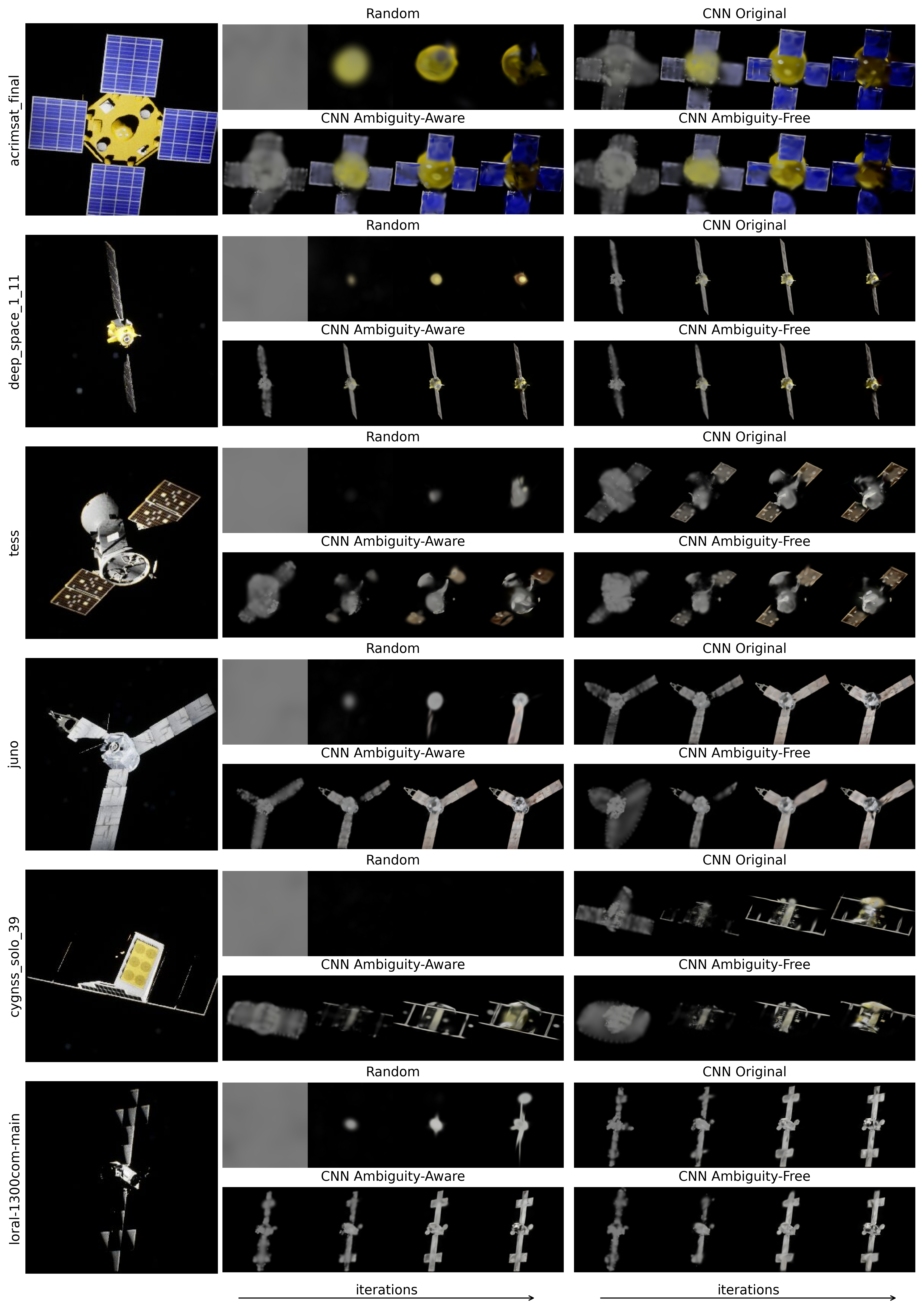}
    \caption{Six example images of satellites (top four from training set and bottom two from test set) from the SPE3R dataset and the accompanying renderings from 3DGS of this image for the 4 initialization styles over the course of the 1,500 training iterations. These trainings are performed using the reference truth poses. }
    \label{fig:gt_pose_comparison_renderings}
\end{figure}

\begin{table}[ht]
    \centering
    \footnotesize
    \caption{Training, evaluation, and timing metrics ($\mu \pm \sigma$) for the SPE3R test set using reference truth poses.}
    \begin{tabular}{r|c c c c}
        \toprule
        \makecell[r]{Init. Style / \\ Metric} & Random & \makecell{CNN \\ Original} & \makecell{CNN \\ Ambiguity Aware} & \makecell{CNN \\ Ambiguity Free} \\
        \hline
        $\mathcal{L}  \; (\downarrow)$ & $0.095 \pm 0.040$ & $\mathbf{0.083 \pm 0.046}$ & $0.084 \pm 0.046$ & $0.084 \pm 0.046$ \\
        $\mathcal{L}_1  \; (\downarrow)$ & $0.065 \pm 0.031$ & $\mathbf{0.054 \pm 0.032}$ & $0.054 \pm 0.032$ & $0.054 \pm 0.032$ \\
        SSIM $(\uparrow)$ & $0.785 \pm 0.078$ & $\mathbf{0.799 \pm 0.103}$ & $0.797 \pm 0.102$ & $0.798 \pm 0.101$ \\
        \hline
        PSNR $(\uparrow)$ & $15.629 \pm 2.405$ & $\mathbf{17.566 \pm 3.616}$ & $17.542 \pm 3.586$ & $17.483 \pm 3.589$ \\
        LPIPS $(\downarrow)$ & $0.403 \pm 0.128$ & $\mathbf{0.276 \pm 0.148}$ & $0.288 \pm 0.149$ & $0.293 \pm 0.144$ \\
        CD $(\downarrow)$ & $1.805 \pm 1.943$ & $\mathbf{0.004 \pm 0.004}$ & $0.006 \pm 0.006$ & $0.006 \pm 0.005$ \\
        \hline
        Init. Time [s] $(\downarrow)$ & $\mathbf{0.005 \pm 0.000}$ & $0.335 \pm 0.021$ & $0.302 \pm 0.062$ & $0.320 \pm 0.044$ \\
        \noalign{\vskip -1.75mm} \\
        2.0 $\times$ LPIPS\textsubscript{best} \makecell[r]{Time to [s] \\ Iters to [iters]} $(\downarrow)$ & \makecell[c]{$3.464 \pm 4.333$ \\ $400 \pm 519$} & \makecell[c]{$\mathbf{0.714 \pm 0.154}$ \\ $\mathbf{39 \pm 18}$} & \makecell[c]{$0.814 \pm 0.405$ \\ $57 \pm 46$} & \makecell[c]{$0.962 \pm 0.750$ \\ $71 \pm 85$} \\
        \noalign{\vskip -1.75mm} \\
        1.5 $\times$ LPIPS\textsubscript{best} \makecell[r]{Time to [s] \\ Iters to [iters]} $(\downarrow)$ & \makecell[c]{$7.847 \pm 5.754$ \\ $892 \pm 661$} & \makecell[c]{$\mathbf{2.335 \pm 2.059}$ \\ $\mathbf{243 \pm 254}$} & \makecell[c]{$3.017 \pm 2.268$ \\ $321 \pm 265$} & \makecell[c]{$3.160 \pm 2.541$ \\ $332 \pm 290$} \\
        \noalign{\vskip -1.75mm} \\
        1.1 $\times$ LPIPS\textsubscript{best} \makecell[r]{Time to [s] \\ Iters to [iters]} $(\downarrow)$ & \makecell[c]{$13.090 \pm 0.330$ \\ $1500 \pm 0$} & \makecell[c]{$\mathbf{8.525 \pm 3.141}$ \\ $\mathbf{971 \pm 334}$} & \makecell[c]{$10.207 \pm 1.620$ \\ $1168 \pm 180$} & \makecell[c]{$11.592 \pm 3.154$ \\ $1318 \pm 323$} \\
        \bottomrule
    \end{tabular}
    \label{tab:gt_pose_test_set_stats}
\end{table}

\autoref{fig:gt_pose_comparison_renderings} shows 6 example satellites and the respective tests from each initialization style. 
The figure is split into 3 columns, with the left-most column showing a reference truth image that the model will try to re-create. 
For each initialization style, renders are shown as the number of training iterations increases (from 0 to 1,500) showing the improvements on the reconstruction quality during training. 

Next, the same trainings are performed, but this time the estimated poses are used. 
For the random initialization, three tests are performed for every satellite where each test uses a different CNN version as the source of pose. 
The statistics of the metrics for the random initialization using poses from the three CNN variants can be seen in Tables \ref{tab:est_pose_randominit_train_set_stats} and \ref{tab:est_pose_randominit_test_set_stats}. 
The same statistics for the CNN initializations are found in Tables \ref{tab:est_pose_sqcnninit_train_set_stats} and \ref{tab:est_pose_sqcnninit_test_set_stats}. 
\autoref{fig:est_pose_comparison_renderings} once again compares the renderings of the reference truth image from 3DGS during training; this time using estimated poses from the CNNs.
The randomly initialized 3DGS model uses the pose estimates from the ambiguity-free version of the CNN.

The pose errors from the three different CNN versions are shown quantitatively in \autoref{tab:pose_errors} and qualitatively in \autoref{fig:pose_estimates_comparison}.
In the latter, the single-shot mesh estimate from each CNN version is projected onto the image plane by the rotation estimate provided by the CNN giving a sense of the nature of the CNN pose errors.

\subsection{Discussion}
In the cases where reference truth poses are used, the initialization techniques implemented in this work out-perform the random initialization as shown in Tables \ref{tab:gt_pose_train_set_stats} and \ref{tab:gt_pose_test_set_stats}.
The timing metrics demonstrate that, although there is a cost to initializing using the CNNs, it is quickly overcome with a faster convergence to a better representation of the target. 
Even when accounting for the time required to initialize, on average the CNN initialization is twice as fast at arriving at an LPIPS score which is only 1.5 times larger than the best received for that satellite.

\begin{table}[!ht]
    \centering
    \footnotesize
    \caption{Training, evaluation, and timing metrics ($\mu \pm \sigma$) for the SPE3R training set using estimated poses from the CNNs and initialized randomly.}
    \begin{tabular}{r|c c c c}
        \toprule
        \makecell[r]{Init. Style \& \\ CNN for Pose / \\ Metric} & \makecell{Random \& \\ Original} & \makecell{Random \& \\ Ambiguity Aware} & \makecell{Random \& \\ Ambiguity Free} \\
        \hline
        $\mathcal{L}  \; (\downarrow)$ & $0.096 \pm 0.050$ & $0.097 \pm 0.051$ & $\mathbf{0.095 \pm 0.050}$ \\
        $\mathcal{L}_1  \; (\downarrow)$ & $0.064 \pm 0.035$ & $0.064 \pm 0.035$ & $\mathbf{0.062 \pm 0.035}$ \\
        SSIM $(\uparrow)$ & $\mathbf{0.774 \pm 0.112}$ & $0.773 \pm 0.115$ & $0.772 \pm 0.112$ \\
        \hline
        PSNR $(\uparrow)$ & $15.842 \pm 2.655$ & $15.867 \pm 2.706$ & $\mathbf{16.227 \pm 2.655}$ \\
        LPIPS $(\downarrow)$ & $0.431 \pm 0.098$ & $0.430 \pm 0.100$ & $\mathbf{0.406 \pm 0.105}$ \\
        CD $(\downarrow)$ & $0.823 \pm 1.424$ & $0.918 \pm 1.540$ & $\mathbf{0.455 \pm 0.826}$ \\
        \hline
        Init. Time [s] $(\downarrow)$ & $0.012 \pm 0.006$ & $0.011 \pm 0.007$ & $\mathbf{0.011 \pm 0.004}$ \\
        \noalign{\vskip -1.75mm} \\
        2.0 $\times$ LPIPS\textsubscript{best} \makecell[r]{Time to [s] \\ Iters to [iters]} $(\downarrow)$ & \makecell[c]{$\mathbf{1.477 \pm 3.270}$ \\ $\mathbf{94 \pm 237}$} & \makecell[c]{$1.561 \pm 3.781$ \\ $101 \pm 275$} & \makecell[c]{$97.844 \pm 199.130$ \\ $77 \pm 144$} \\
        \noalign{\vskip -1.75mm} \\
        1.5 $\times$ LPIPS\textsubscript{best} \makecell[r]{Time to [s] \\ Iters to [iters]} $(\downarrow)$ & \makecell[c]{$7.162 \pm 8.189$ \\ $504 \pm 587$} & \makecell[c]{$\mathbf{6.995 \pm 8.068}$ \\ $\mathbf{498 \pm 584}$} & \makecell[c]{$559.667 \pm 648.110$ \\ $426 \pm 489$} \\
        \noalign{\vskip -1.75mm} \\
        1.1 $\times$ LPIPS\textsubscript{best} \makecell[r]{Time to [s] \\ Iters to [iters]} $(\downarrow)$ & \makecell[c]{$19.052 \pm 5.167$ \\ $1369 \pm 369$} & \makecell[c]{$\mathbf{18.763 \pm 5.202}$ \\ $\mathbf{1355 \pm 374}$} & \makecell[c]{$1863.489 \pm 615.575$ \\ $1325 \pm 384$} \\
        \bottomrule
    \end{tabular}
    \label{tab:est_pose_randominit_train_set_stats}
\end{table}

\begin{table}[!ht]
    \centering
    \footnotesize
    \caption{Training, evaluation, and timing metrics ($\mu \pm \sigma$) for the SPE3R training set using estimated poses from the CNNs and initialized using the CNN.}
    \begin{tabular}{r|c c c c}
        \toprule
        \makecell[r]{Init. Style \& \\ CNN for Pose / \\ Metric} & \makecell{CNN \& \\ Original} & \makecell{CNN \& \\ Ambiguity Aware} & \makecell{CNN \& \\ Ambiguity Free} \\
        \hline
        $\mathcal{L}  \; (\downarrow)$ & $0.096 \pm 0.048$ & $0.096 \pm 0.049$ & $\mathbf{0.094 \pm 0.049}$ \\
        $\mathcal{L}_1  \; (\downarrow)$ & $0.064 \pm 0.035$ & $0.064 \pm 0.035$ & $\mathbf{0.061 \pm 0.035}$ \\
        SSIM $(\uparrow)$ & $\mathbf{0.776 \pm 0.104}$ & $0.776 \pm 0.106$ & $0.770 \pm 0.108$ \\
        \hline
        PSNR $(\uparrow)$ & $15.966 \pm 2.730$ & $15.951 \pm 2.722$ & $\mathbf{16.448 \pm 2.691}$ \\
        LPIPS $(\downarrow)$ & $0.412 \pm 0.105$ & $0.413 \pm 0.104$ & $\mathbf{0.354 \pm 0.127}$ \\
        CD $(\downarrow)$ & $\mathbf{0.014 \pm 0.010}$ & $0.016 \pm 0.015$ & $0.028 \pm 0.018$ \\
        \hline
        Init. Time [s] $(\downarrow)$ & $0.187 \pm 0.014$ & $0.188 \pm 0.011$ & $\mathbf{0.183 \pm 0.009}$ \\
        \noalign{\vskip -1.75mm} \\
        2.0 $\times$ LPIPS\textsubscript{best} \makecell[r]{Time to [s] \\ Iters to [iters]} $(\downarrow)$ & \makecell[c]{$0.636 \pm 0.411$ \\ $29 \pm 26$} & \makecell[c]{$\mathbf{0.587 \pm 0.093}$ \\ $\mathbf{26 \pm 7}$} & \makecell[c]{$30.152 \pm 4.983$ \\ $25 \pm 0$} \\
        \noalign{\vskip -1.75mm} \\
        1.5 $\times$ LPIPS\textsubscript{best} \makecell[r]{Time to [s] \\ Iters to [iters]} $(\downarrow)$ & \makecell[c]{$\mathbf{2.549 \pm 5.599}$ \\ $\mathbf{160 \pm 380}$} & \makecell[c]{$2.979 \pm 6.311$ \\ $189 \pm 429$} & \makecell[c]{$32.436 \pm 12.881$ \\ $27 \pm 8$} \\
        \noalign{\vskip -1.75mm} \\
        1.1 $\times$ LPIPS\textsubscript{best} \makecell[r]{Time to [s] \\ Iters to [iters]} $(\downarrow)$ & \makecell[c]{$\mathbf{15.758 \pm 9.360}$ \\ $\mathbf{1069 \pm 640}$} & \makecell[c]{$16.618 \pm 8.549$ \\ $1131 \pm 584$} & \makecell[c]{$567.877 \pm 579.787$ \\ $402 \pm 389$} \\
        \bottomrule
    \end{tabular}
    \label{tab:est_pose_sqcnninit_train_set_stats}
\end{table}

From \autoref{fig:gt_pose_comparison_renderings} it is clear that for the satellites from the training set (top 4), the CNNs estimate good initial 3D models which kick-start the 3DGS training.
In some cases, such as with the \textit{acrimsat\_final} satellite, this allows the 3DGS model to learn the structure of the three additional solar panels which the random initialization is not capable of learning. 
Poor initializations still outperform cases with random initialization, as is seen with the test set satellites (bottom 2) and especially with \textit{cygnss\_solo\_39}.
Here, the original CNN has a completely incorrect initial estimate, but the 3DGS training is able to overcome this poor estimate and return a refined 3D model of the target. 
The random initialization, on the other hand, is not able to learn this satellite well enough to reconstruct this particular image, likely due to the poor lighting conditions.

\begin{figure}[!p]
    \centering
    \includegraphics[width=\linewidth]{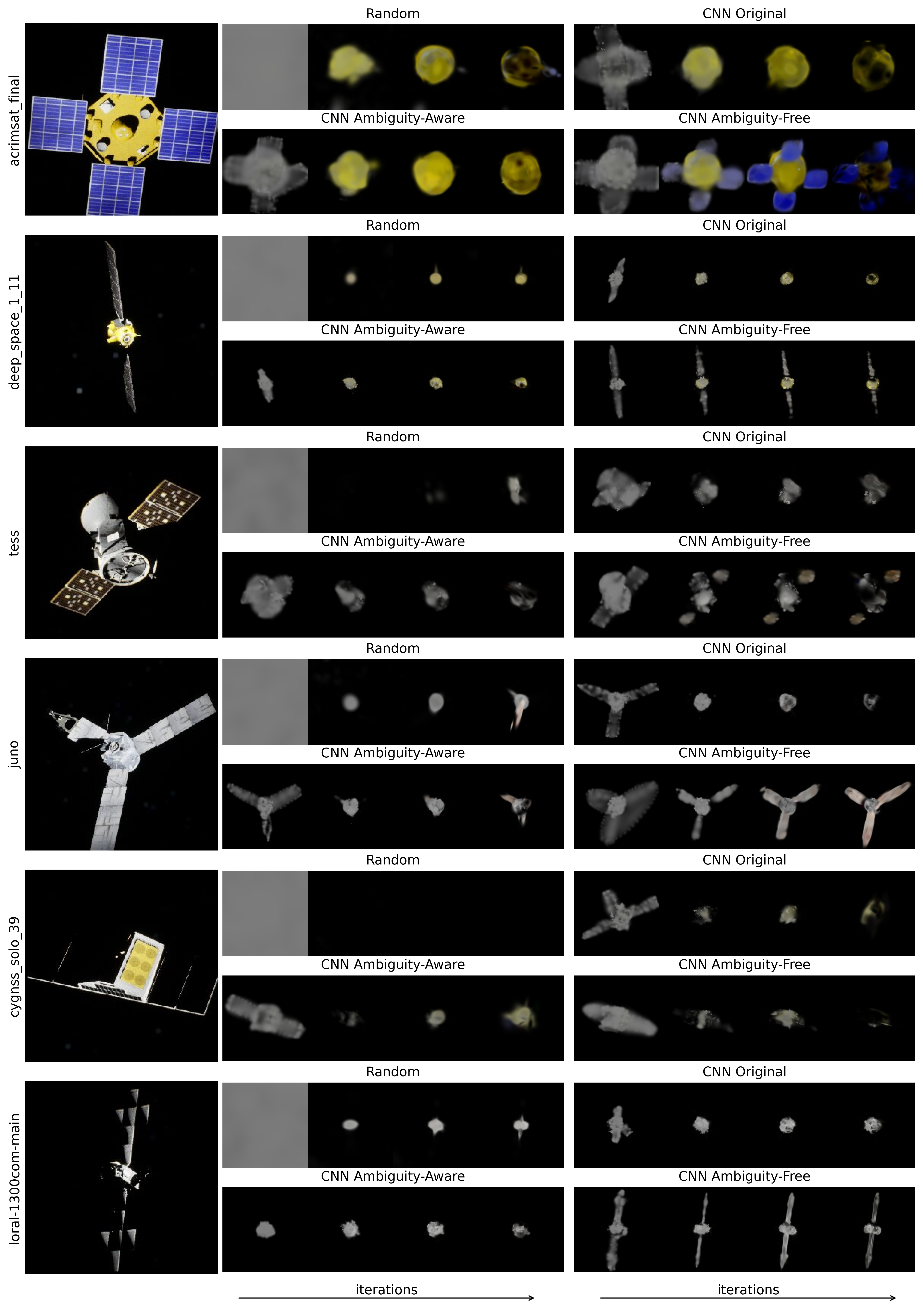}
    \caption{Six example images of satellites (top four from training set and bottom two from test set) from the SPE3R dataset and the accompanying renderings from 3DGS of this image for the 4 initialization styles over the course of the 1,500 training iterations. These trainings are performed using the estimated poses. The random initialization is trained using the poses from the ambiguity-free CNN. }
    \label{fig:est_pose_comparison_renderings}
\end{figure}

\begin{figure}[p]
    \centering
    \includegraphics[width=\linewidth]{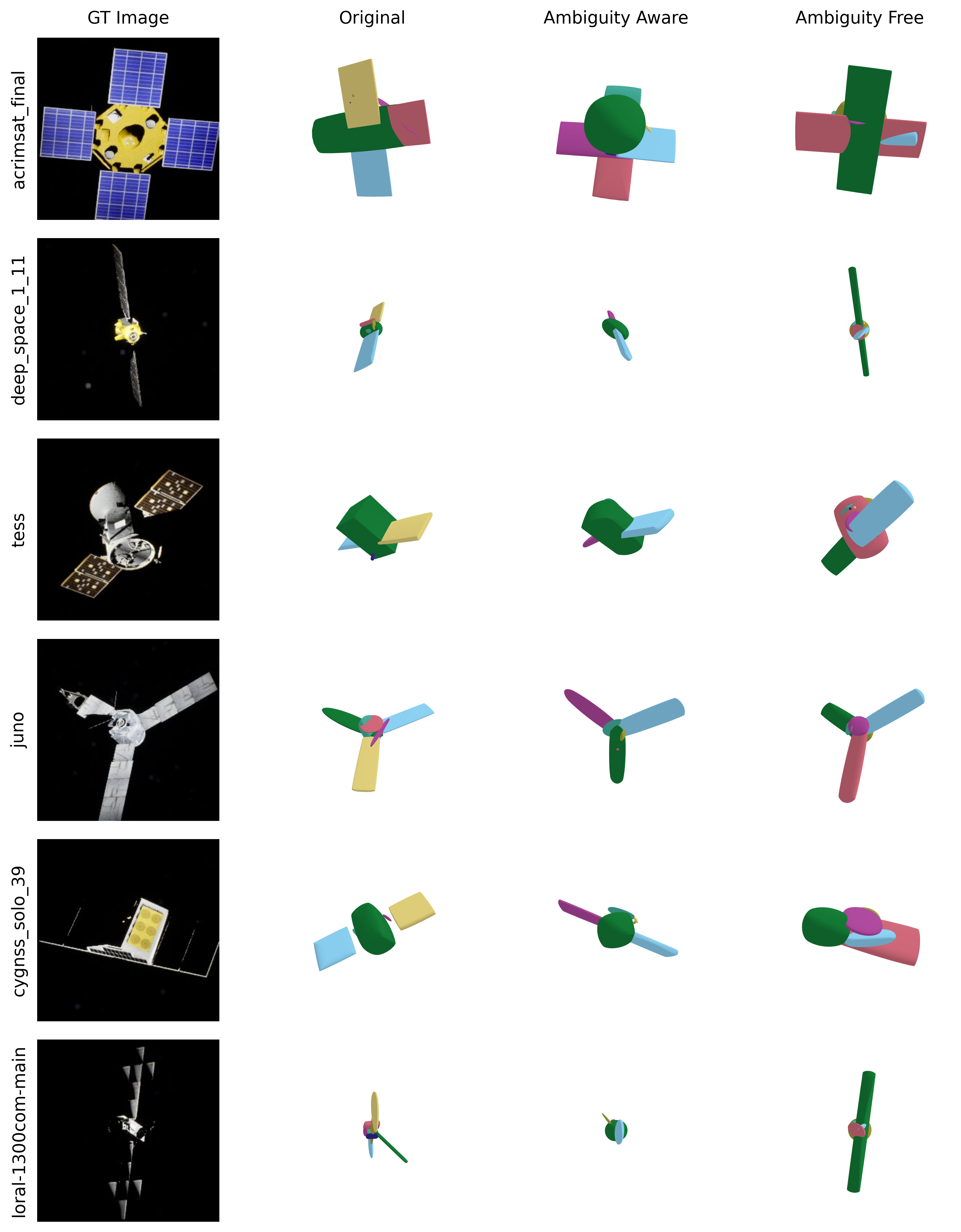}
    \caption{Six example images of satellites (top four from training set and bottom two from test set) from the SPE3R dataset and the accompanying single-shot mesh estimate from the CNNs rotated by their rotation estimate.}
    \label{fig:pose_estimates_comparison}
\end{figure}

\begin{table}[!ht]
    \centering
    \footnotesize
    \caption{Training, evaluation, and timing metrics ($\mu \pm \sigma$) for the SPE3R test set using estimated poses from the CNNs and initialized randomly.}
    \begin{tabular}{r|c c c c}
        \toprule
        \makecell[r]{Init. Style \& \\ CNN for Pose / \\ Metric} & \makecell{Random \& \\ Original} & \makecell{Random \& \\ Ambiguity Aware} & \makecell{Random \& \\ Ambiguity Free} \\
        \hline
        $\mathcal{L}  \; (\downarrow)$ & $0.089 \pm 0.044$ & $0.089 \pm 0.044$ & $\mathbf{0.089 \pm 0.044}$ \\
        $\mathcal{L}_1  \; (\downarrow)$ & $0.061 \pm 0.034$ & $0.061 \pm 0.034$ & $\mathbf{0.060 \pm 0.033}$ \\
        SSIM $(\uparrow)$ & $\mathbf{0.798 \pm 0.088}$ & $0.796 \pm 0.086$ & $0.795 \pm 0.086$ \\
        \hline
        PSNR $(\uparrow)$ & $16.178 \pm 3.659$ & $16.185 \pm 3.656$ & $\mathbf{16.372 \pm 3.584}$ \\
        LPIPS $(\downarrow)$ & $0.423 \pm 0.103$ & $0.423 \pm 0.102$ & $\mathbf{0.409 \pm 0.112}$ \\
        CD $(\downarrow)$ & $1.625 \pm 1.202$ & $2.212 \pm 1.040$ & $\mathbf{1.229 \pm 1.022}$ \\
        \hline
        Init. Time [s] $(\downarrow)$ & $0.013 \pm 0.005$ & $0.013 \pm 0.008$ & $\mathbf{0.008 \pm 0.004}$ \\
        \noalign{\vskip -1.75mm} \\
        2.0 $\times$ LPIPS\textsubscript{best} \makecell[r]{Time to [s] \\ Iters to [iters]} $(\downarrow)$ & \makecell[c]{$1.292 \pm 1.182$ \\ $57 \pm 51$} & \makecell[c]{$\mathbf{1.208 \pm 1.000}$ \\ $\mathbf{54 \pm 45}$} & \makecell[c]{$68.517 \pm 67.029$ \\ $57 \pm 51$} \\
        \noalign{\vskip -1.75mm} \\
        1.5 $\times$ LPIPS\textsubscript{best} \makecell[r]{Time to [s] \\ Iters to [iters]} $(\downarrow)$ & \makecell[c]{$10.275 \pm 14.165$ \\ $486 \pm 646$} & \makecell[c]{$\mathbf{10.273 \pm 14.105}$ \\ $\mathbf{486 \pm 646}$} & \makecell[c]{$807.581 \pm 1088.077$ \\ $496 \pm 642$} \\
        \noalign{\vskip -1.75mm} \\
        1.1 $\times$ LPIPS\textsubscript{best} \makecell[r]{Time to [s] \\ Iters to [iters]} $(\downarrow)$ & \makecell[c]{$20.028 \pm 10.396$ \\ $1257 \pm 509$} & \makecell[c]{$\mathbf{19.905 \pm 10.353}$ \\ $\mathbf{1257 \pm 509}$} & \makecell[c]{$1893.214 \pm 841.195$ \\ $1182 \pm 514$} \\
        \bottomrule
    \end{tabular}
    \label{tab:est_pose_randominit_test_set_stats}
\end{table}

\begin{table}[!ht]
    \centering
    \footnotesize
    \caption{Training, evaluation, and timing metrics ($\mu \pm \sigma$) for the SPE3R test set using estimated poses from the CNNs and initialized using the CNN.}
    \begin{tabular}{r|c c c c}
        \toprule
        \makecell[r]{Init. Style \& \\ CNN for Pose / \\ Metric} & \makecell{CNN \& \\ Original} & \makecell{CNN \& \\ Ambiguity Aware} & \makecell{CNN \& \\ Ambiguity Free} \\
        \hline
        $\mathcal{L}  \; (\downarrow)$ & $\mathbf{0.089 \pm 0.043}$ & $0.089 \pm 0.044$ & $0.089 \pm 0.044$ \\
        $\mathcal{L}_1  \; (\downarrow)$ & $0.060 \pm 0.033$ & $0.060 \pm 0.033$ & $\mathbf{0.058 \pm 0.032}$ \\
        SSIM $(\uparrow)$ & $\mathbf{0.799 \pm 0.083}$ & $0.795 \pm 0.087$ & $0.788 \pm 0.093$ \\
        \hline
        PSNR $(\uparrow)$ & $16.356 \pm 3.872$ & $16.405 \pm 3.818$ & $\mathbf{16.632 \pm 3.435}$ \\
        LPIPS $(\downarrow)$ & $0.403 \pm 0.110$ & $0.404 \pm 0.115$ & $\mathbf{0.357 \pm 0.130}$ \\
        CD $(\downarrow)$ & $\mathbf{0.030 \pm 0.018}$ & $0.031 \pm 0.018$ & $0.035 \pm 0.021$ \\
        \hline
        Init. Time [s] $(\downarrow)$ & $0.210 \pm 0.042$ & $0.199 \pm 0.028$ & $\mathbf{0.181 \pm 0.005}$ \\
        \noalign{\vskip -1.75mm} \\
        2.0 $\times$ LPIPS\textsubscript{best} \makecell[r]{Time to [s] \\ Iters to [iters]} $(\downarrow)$ & \makecell[c]{$0.717 \pm 0.193$ \\ $25 \pm 0$} & \makecell[c]{$\mathbf{0.675 \pm 0.180}$ \\ $\mathbf{25 \pm 0}$} & \makecell[c]{$34.106 \pm 2.224$ \\ $25 \pm 0$} \\
        \noalign{\vskip -1.75mm} \\
        1.5 $\times$ LPIPS\textsubscript{best} \makecell[r]{Time to [s] \\ Iters to [iters]} $(\downarrow)$ & \makecell[c]{$1.348 \pm 1.411$ \\ $54 \pm 60$} & \makecell[c]{$\mathbf{1.310 \pm 1.412}$ \\ $\mathbf{54 \pm 60}$} & \makecell[c]{$34.106 \pm 2.224$ \\ $25 \pm 0$} \\
        \noalign{\vskip -1.75mm} \\
        1.1 $\times$ LPIPS\textsubscript{best} \makecell[r]{Time to [s] \\ Iters to [iters]} $(\downarrow)$ & \makecell[c]{$\mathbf{14.284 \pm 14.152}$ \\ $\mathbf{711 \pm 684}$} & \makecell[c]{$16.240 \pm 13.471$ \\ $896 \pm 699$} & \makecell[c]{$401.397 \pm 330.459$ \\ $239 \pm 195$} \\
        \bottomrule
    \end{tabular}
    \label{tab:est_pose_sqcnninit_test_set_stats}
\end{table}

From Tables \ref{tab:est_pose_randominit_train_set_stats}, \ref{tab:est_pose_randominit_test_set_stats}, \ref{tab:est_pose_sqcnninit_train_set_stats}, and \ref{tab:est_pose_sqcnninit_test_set_stats} when reference-truth poses are replaced with estimates poses, initialization from the CNN continues to lead to better and faster reconstructions compared to the baseline method. 
However, the cost of the point-cloud alignment estimation method used in this work also becomes clear in these results. 
The time required to reach the landmark LPIPS levels is now multiple orders of magnitude larger purely due to the cost of the point-cloud alignment method implemented in this work.

The metrics themselves are worse using estimated poses than they were when the training used reference truth values; the true effect of this trend is best appreciated in \autoref{fig:est_pose_comparison_renderings}. 
Here it becomes clear that only the ambiguity-free CNN provides sufficiently accurate pose estimates to achieve good reconstructions. 
Because of this, the random initialization shown in \autoref{fig:est_pose_comparison_renderings} is trained using poses from post-processing the ambiguity-free CNN's estimates with point cloud alignment. 
However, even with the best pose estimates, the random initialization is frequently unable to reconstruct the solar panels.
The CNN initialization with the ambiguity-free poses overcomes the pose errors and is able to reconstruct the 3D model including the solar panels, although worse than when using reference truth poses. 
The exception to this trend is the \textit{cygnss\_solo\_39} spacecraft, which the 3DGS model fails to reconstruct regardless of initialization method. 
This was the most challenging spacecraft when using reference truth poses, so it is of little surprise that when faced with noisy pose estimates it would fail.

The actual pose errors in \autoref{tab:pose_errors} show that the ambiguity-free version of the CNN does result in the lowest pose error. 
This error is still quite high, and intuitively should cause issues when training 3DGS; however, the reason it is often still successful in reconstructing the true 3D structure is best exemplified by \autoref{fig:pose_estimates_comparison}.
Here is it clear that while the ambiguity-free version of the CNN is not error-free, the rotation errors are about the body axis aligned with the solar panels. 
This means that although there is error, the estimate places the solar panels in the correct approximate location in the image, allowing the 3DGS model to properly learn their location. 
The other CNN versions exhibit random errors, which means that the solar panels are not in the correct location, and this can lead to them not being learned by 3DGS as they cancel out when averaged over many training images.

\begin{table}[ht]
    \centering
    \footnotesize
    \caption{Pose errors from the different CNN versions on the two SPE3R dataset splits.}
    \begin{tabular}{r | c c | c c}
        \toprule
        & \multicolumn{2}{c|}{Train}  & \multicolumn{2}{c}{Test} \\
        Metric / CNN Version & eR $[^\circ]$ & eT & eR $[^\circ]$ & eT  \\
        \hline
        Original & $62.632 \pm 28.657$ & $\mathbf{0.054 \pm 0.048}$ & $70.614 \pm 22.213$ & $0.099 \pm 0.075$ \\
        Ambiguity-Aware & $62.880 \pm 28.449$ & $0.073 \pm 0.058$ & $69.497 \pm 23.450$ & $0.103 \pm 0.081$ \\
        Ambiguity-Free & $\mathbf{46.797 \pm 32.190}$ & $0.070 \pm 0.059$ & $\mathbf{63.554 \pm 25.966}$ & $\mathbf{0.092 \pm 0.070}$ \\
        \bottomrule
    \end{tabular}
    \label{tab:pose_errors}
\end{table}

\section{Conclusion}
This work addressed two key challenges in implementing novel view synthesis for space applications, namely the computational cost and the availability of pose priors. 
Building on prior work, this work leverages fast one-shot coarse shape estimates from a Convolutional Neural Network (CNN) in order to reduce the training time of 3D Gaussian Splatting (3DGS), and in some cases even significantly improve the reconstruction.
This work also attempts to use the pose estimates -- both explicit and implicit -- available from the CNN variants, and demonstrates their effects on the reconstruction of the model with 3DGS. 
The results demonstrate that the additional time spent initializing 3DGS via CNN inference is time well spent, as the CNN-initialized 3DGS models quickly catch up in reconstruction quality and are able to arrive at refined 3D models quicker than the random initialization.
This is even true for models the CNN did not see during training, which sometimes result in poor estimates of the true spacecraft shape, as the 3DGS training is able to overcome the poor initialization and still outperform the baseline. 
In the case of training 3DGS using estimated poses, it is clear that the ambiguity-free version of the network is most successful as it is capable of providing pose estimates which at least align the solar panels along the correct principal direction.
The other CNN versions provide pose estimates that are too noisy, leading to the solar panels disappearing from the model during 3DGS training. 

Further fine-tuning of the hyper-parameters, both of 3DGS and of the CNNs could lead to better initialization and faster and more precise reconstructions.
Implementations of this methodology in space-grade processors, and testing on space -- or space-like -- imagery would also allow for their evaluation for space-applicability.
This work clearly demonstrates the need for precise pose estimates in training 3DGS which requires further research in the case of previously unknown targets. 
Additional research could also investigate the direct refinement of the superquadric primitives, replacing the 3D Gaussians from the original literature and providing a more direct initialization process that maintains the explainability of the Gaussians. 
\hr{Finally, a strong limitation in implementing the ambiguity-free CNN to do pose estimation in space is the time required to perform point-cloud alignment.
This should be reduced by the use of sequential data streams and pose filtering to arrive at good initial orientations that can be refined, but further research is required to implement and demonstrate that this does not become compute prohibitive.}

\section{Acknowledgments}
This work was supported by Redwire Space Corporation through Stanford’s Center for AEroSpace Autonomy Research (CAESAR).

\bibliographystyle{AAS_publication}   
\bibliography{references}   

\end{document}